\documentclass[10pt,twocolumn,letterpaper]{article}
\usepackage[accsupp]{axessibility}  %
\usepackage{iccv}              %

\definecolor{iccvblue}{rgb}{0.21,0.49,0.74}
\usepackage[pagebackref,breaklinks,colorlinks,allcolors=iccvblue]{hyperref}

\def\paperID{2055} %
\def\confName{ICCV}
\def\confYear{2025}

\usepackage{times}
\usepackage{epsfig}
\usepackage{graphicx}
\usepackage{amsmath}
\usepackage{amssymb}
\usepackage{color,xcolor}

\usepackage{comment}

\usepackage{pifont}

\usepackage{microtype}
\usepackage[font=small]{caption}
\usepackage{arydshln}
\usepackage{tabularx}
\usepackage{adjustbox}
\usepackage{array}
\usepackage{booktabs}
\usepackage{colortbl}
\usepackage{float,wrapfig}
\usepackage{hhline}
\usepackage{multirow}
\usepackage{subcaption} %
\usepackage[percent]{overpic}

\usepackage{stmaryrd}

\usepackage{amsmath,amsfonts,amssymb}
\usepackage{bm}
\usepackage{nicefrac}
\usepackage{microtype}
\usepackage{dsfont}
\usepackage{changepage}
\usepackage{extramarks}
\usepackage{fancyhdr}
\usepackage{setspace}
\usepackage{soul}
\usepackage{xspace}

\usepackage{algorithm, algpseudocode}
\usepackage{enumitem}
\usepackage[title]{appendix}

\usepackage{duckuments}
\usepackage{subcaption}

\usepackage{animate}
\usepackage{tabularx}

\usepackage[multiple]{footmisc}
\usepackage{bigfoot}
\DeclareNewFootnote{AAffil}[arabic]
\DeclareNewFootnote{ANote}[fnsymbol]

\usepackage{amsmath}
\usepackage{amssymb}
\usepackage{booktabs}
\usepackage{cuted}
\usepackage{capt-of}
\usepackage{graphicx}
\usepackage{xcolor}
\usepackage{comment}
\usepackage{multirow}
\usepackage{bm}

\usepackage{fp}
\usepackage{expl3}[2012-07-08]
\ExplSyntaxOn
\ExplSyntaxOff

\usepackage{amsmath,amsfonts,bm}

\def\x{{x}}

\def\xi{{\x_i}}

\newcommand{\reffig}[1]{Figure~\ref{fig:#1}}
\newcommand{\refsec}[1]{Section~\ref{sec:#1}}

\newcommand{\reftbl}[1]{Table~\ref{tbl:#1}}
\newcommand{\refalg}[1]{Algorithm~\ref{alg:#1}}

\newcommand{\lblfig}[1]{\label{fig:#1}}
\newcommand{\lblsec}[1]{\label{sec:#1}}

\newcommand{\lbltbl}[1]{\label{tbl:#1}}
\newcommand{\lblalg}[1]{\label{alg:#1}}

\newcommand{\ignorethis}[1]{}

\newcommand{\myparagraph}[1]{\vspace{0pt} \smallskip \noindent \textbf{#1}}

\def\eqref#1{equation~\ref{#1}}

\def\1{\bm{1}}

\def\rvb{{\mathbf{b}}}

\def\rvn{{\mathbf{n}}}

\def\rvp{{\mathbf{p}}}

\def\rvx{{\mathbf{x}}}

\def\rmA{{\mathbf{A}}}

\def\rmO{{\mathbf{O}}}
\def\rmP{{\mathbf{P}}}
\def\rmQ{{\mathbf{Q}}}

\def\rmS{{\mathbf{S}}}

\DeclareMathAlphabet{\mathsfit}{\encodingdefault}{\sfdefault}{m}{sl}
\SetMathAlphabet{\mathsfit}{bold}{\encodingdefault}{\sfdefault}{bx}{n}

\def\gE{{\mathcal{E}}}

\def\gO{{\mathcal{O}}}

\def\gV{{\mathcal{V}}}

\newcommand{\ignore}[1]{}

\makeatletter
\DeclareRobustCommand\onedot{\futurelet\@let@token\@onedot}
\def\@onedot{\ifx\@let@token.\else.\null\fi\xspace}

\makeatother

\definecolor{MyDarkBlue}{rgb}{0,0.08,1}
\definecolor{MyDarkGreen}{rgb}{0.02,0.6,0.02}
\definecolor{MyDarkRed}{rgb}{0.8,0.02,0.02}
\definecolor{MyDarkOrange}{rgb}{0.40,0.2,0.02}
\definecolor{MyPurple}{RGB}{111,0,255}
\definecolor{MyRed}{rgb}{1.0,0.0,0.0}
\definecolor{MyGold}{rgb}{0.75,0.6,0.12}
\definecolor{MyDarkgray}{rgb}{0.66, 0.66, 0.66}
\definecolor{myorange}{RGB}{255,69,0}
\definecolor{derekblue}{RGB}{144,210,236}

\definecolor{revisionColor}{RGB}{0,0,0}

\newcommand{\ours}{OAT\xspace}
\newcommand{\ourAR}{OctreeGPT\xspace}

\title{Efficient Autoregressive Shape Generation via \\  Octree-Based Adaptive Tokenization}

\author{Kangle Deng\textsuperscript{1,2}\thanks{Work done when interning at Roblox.} \quad Hsueh-Ti Derek Liu\textsuperscript{2}  \quad Yiheng Zhu\textsuperscript{2} \quad Xiaoxia Sun\textsuperscript{2} \quad Chong Shang\textsuperscript{2} \\
\quad Kiran S. Bhat\textsuperscript{2} 
\quad Deva Ramanan\textsuperscript{1} 
\quad Jun-Yan Zhu\textsuperscript{1} \quad Maneesh Agrawala\textsuperscript{2,3} 
\quad Tinghui Zhou\textsuperscript{2}
\\
\textsuperscript{1}Carnegie Mellon University \qquad \textsuperscript{2}Roblox \qquad \textsuperscript{3}Stanford University }

\begin{document}

\maketitle
\begin{strip}\centering
\vspace{-40pt}
\includegraphics[width=\linewidth, trim={0 0cm 0 0cm},clip]{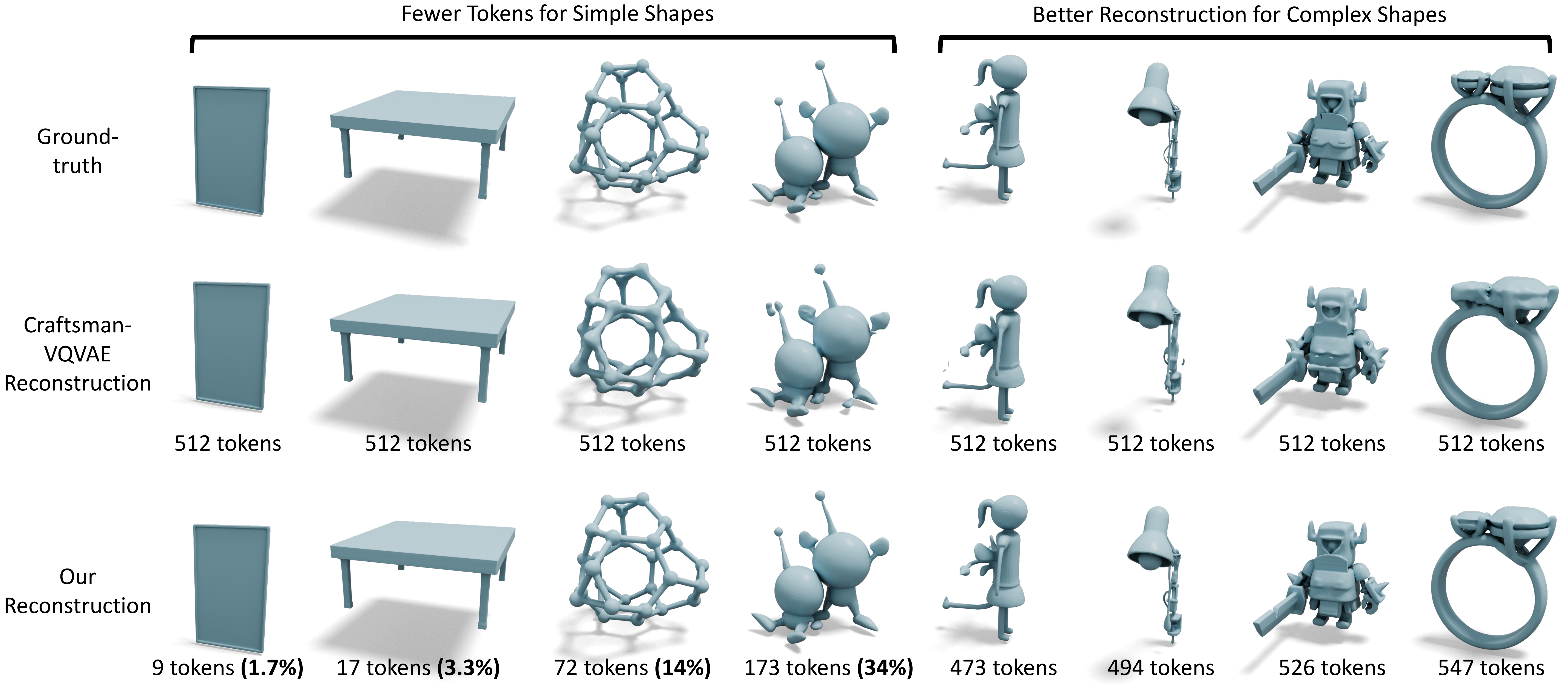}
\vspace{-20pt}
\captionof{figure}{ We propose an Octree-based Adaptive shape Tokenization (\ours) that dynamically allocates tokens based on shape complexity. Our approach achieves {\em better} reconstruction quality with {\em fewer} tokens on average (439 compared to 512 on the full test set) by intelligently distributing more tokens to complex shapes while saving on simpler ones.
\lblfig{teaser}
}
\end{strip}

\begin{abstract}
Many 3D generative models rely on variational autoencoders (VAEs) to learn compact shape representations. 
However, existing methods encode all shapes into a fixed-size token, disregarding the inherent variations in scale and complexity across 3D data. This leads to inefficient latent representations that can compromise downstream generation.
We address this challenge by introducing Octree-based Adaptive Tokenization, a novel framework that adjusts the dimension of latent representations according to shape complexity. Our approach constructs an adaptive octree structure guided by a quadric-error-based subdivision criterion and allocates a shape latent vector to each octree cell using a query-based transformer. Building upon this tokenization, we develop an octree-based autoregressive generative model that effectively leverages these variable-sized representations in shape generation.
Extensive experiments demonstrate that our approach reduces token counts by 50\% compared to fixed-size methods while maintaining comparable visual quality.  
When using a similar token length, our method produces significantly higher-quality shapes. 
When incorporated with our downstream generative model, our method creates more detailed and diverse 3D content than existing approaches.
Please check our project page: \url{https://oat-3d.github.io/}.
\end{abstract}

\section{Introduction}
\lblsec{intro}

Recent advances in generative models have revolutionized the field of 3D content creation, enabling diverse applications, including shape generation~\cite{mescheder2019occupancy,vahdat2022lion,li2024craftsman}, text-to-3D generation~\cite{poole2022dreamfusion,lin2023magic3d,wang2023prolificdreamer}, text-driven mesh texturing~\cite{chen2023text2tex,deng2024flashtex}, single-image 3D generation~\cite{liu2023zero123,long2024wonder3d}, and 3D scene editing~\cite{liu2021editing,haque2023instruct}.
One popular paradigm among state-of-the-art methods employs 3D-native diffusion or autoregressive models~\cite{li2024craftsman,zhao2025hunyuan3d,zhang2024clay,ren2024xcube,xiong2024octfusion} on top of 3D latents learned from large-scale datasets. As a result, the effectiveness of these models heavily depends on how well 3D shapes are represented and encoded as latent representations.

Effective latent representations for 3D shapes must address several fundamental challenges. First, 3D data is inherently sparse, with meaningful information concentrated primarily on surfaces rather than distributed throughout the volume. Second, real-world objects vary in geometric complexity, ranging from simple primitives to intricate structures with fine details, requiring representation structures that can adapt accordingly. Third, the encoding process must take into account capturing fine local details while preserving the global geometric structure.

Most existing shape VAEs~\cite{zhang20233dshape2vecset,li2024craftsman,zhang2024clay,zhao2025hunyuan3d} encode shapes into fixed-size latent representations and fail to adapt to the inherent variations in geometric complexity within such shapes.
As shown in \reffig{teaser} (bottom), objects are encoded with identical latent capacity regardless of their scale, sparsity, or complexity, resulting in inefficient compression and degraded performance in downstream generative models.
While some approaches~\cite{ren2024xcube,xiong2024octfusion} leverage sparse voxel representations like octrees to account for sparsity, they still subdivide any cell containing surface geometry to the finest level, thus failing to adapt to shape complexity.
As illustrated in \reffig{octree_comparison}, a simple cube with only eight vertices requires similar representation capacity as a highly detailed sculpture in traditional octree structures. Ideally, hierarchical shape representations should adapt to the complexity of different regions within a shape. For instance, in the bottom right of \reffig{octree_comparison}, complex structures like a tree canopy should require finer subdivision than simpler regions like the trunk.

\begin{figure}[t]
    \centering
    \includegraphics[width=\linewidth]{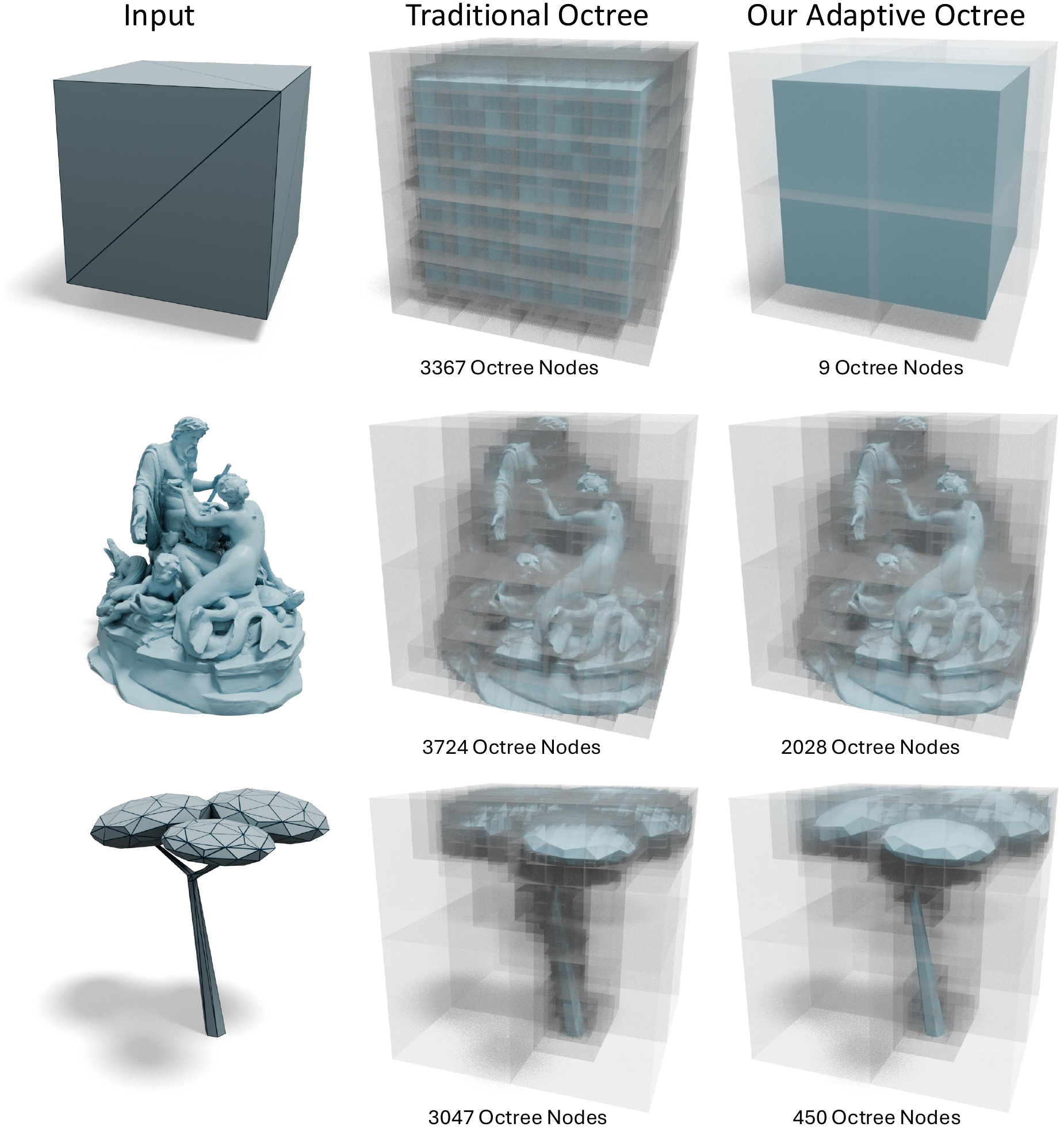}
    \vspace{-20 pt}
    \caption{Traditional octree construction subdivides each octant based on whether the octant contains any mesh element. This construction always subdivides to the maximum depth (set to 6 in this example), leading to a similar amount of nodes for simple (top) and complex (middle) shapes.
    In contrast, our approach terminates subdivision when the local geometry is simple (e.g., a plane), leading to an adaptive octree that better reflects the shape complexity.
    }
    \vspace{-10 pt}
    \lblfig{octree_comparison}
\end{figure}

To address these challenges, we propose an Octree-based Adaptive Tokenization. Our approach dynamically adjusts the latent representation based on local geometric complexity measured by quadric error, efficiently representing both simple and intricate regions with appropriate detail levels. As shown in \reffig{teaser}, our approach achieves better reconstruction quality with comparable or fewer shape tokens. 
By developing an Octree-based autoregressive generative model, we verify that our efficient variable-sized shape tokenization is beneficial to downstream generation tasks. 
Experiments show our generated results are generally better than those of existing baselines regarding FID, KID, and CLIP scores.

\section{Related Work}
\lblsec{related_work}

\myparagraph{3D Generation.} 
Recent 3D generation methods have achieved remarkable results by leveraging pre-trained large-scale 2D diffusion models~\cite{rombach2022high}. Approaches like DreamFusion~\cite{poole2022dreamfusion} and DreamGaussian~\cite{tang2023dreamgaussian} use 2D diffusion priors to optimize 3D representations, such as Neural Radiance Fields~\cite{mildenhall2020nerf} and  Gaussian Splats~\cite{kerbl3Dgaussians}. Subsequent works have improved performance with new loss functions and 3D representations~\cite{wang2023prolificdreamer, ye2025dreamreward, mcallister2024rethinking, tran2025diverse, katzir2023noise, lukoianov2024sdi, lin2023magic3d, Chen_2023fantasia3D, metzer2022latent, sun2023dreamcraft3d, sweetdreamer, long2023wonder3d, michel2022text2mesh}. However, these methods often require extensive iterative optimizations, making them impractical for real-world applications. To reduce inference time, feed-forward methods have been developed that synthesize multi-view consistent images of the same object followed by 3D reconstruction~\cite{hong2023lrm, li2023instant3d, xu2023dmv3d, TripoSR2024,liu2023zero1to3,liu2023one,liu2024one,zhang2024gs,zou2024triplane,tang2024lgm}. Nonetheless, approaches leveraging 2D diffusion priors alone without 3D-native supervision tend to suffer from challenges in modeling refined geometric structures and complex surfaces, especially for shapes of high concavity. 
 
 More recently, a wave of 3D-native generative models~\cite{li2024craftsman, zhang2024clay, zhao2025hunyuan3d, ren2024xcube, xiong2024octfusion, xiang2024trellis, liu2025meshformer} has emerged, aiming to train directly on raw 3D assets rather than relying on 2D diffusion priors. These methods have achieved superior generation quality compared to their predecessors thanks to the 3D-native architecture design. Another line of work explores auto-regressive methods for direct mesh generation with artist-like topology~\cite{tang2024edgerunner, siddiqui2024meshgpt, chen2024meshanything, chen2024meshanythingv2, hao2024meshtron, weng2024scaling}. Due to tokenization inefficiency and challenges in scaling up the context window, these methods are still struggling to model high-poly meshes with complex surfaces. In contrast, our work aims to explore more efficient tokenization schemes that encode shapes into compact yet expressive representations for 3D-native generation.

\myparagraph{Compact 3D latent representations.} Representing 3D shapes with compact latent representations has become increasingly popular in 3D generative modeling. One line of work, pioneered by 3DShape2VecSet~\cite{park2018photoshape}, advocates encoding 3D shapes into latent vector sets that can be decoded into diverse geometry representations such as occupancy fields~\cite{li2024craftsman,zhang2024clay,park2018photoshape,wu2024direct3d,zhao2023michelangelo}, signed distance fields~\cite{chen2024dora,zhao2025hunyuan3d}, and meshes~\cite{tang2024edgerunner}. These methods encode all shapes into a fixed-length vector, and do not adaptively adjust the representation budget based on shape complexity. Other work~\cite{wu2024blockfusion,shue2023triplane_diffusion} learns latent space from triplanes, but achieving high-fidelity triplane representations remains challenging, which limits their accuracy, especially for complex shapes.

An alternative direction focuses on structured 3D latent representations to better leverage the spatial hierarchies inherent in the underlying geometry. For instance, sparse voxel grids coupled with feature-rich latents or attributes, as proposed in XCube~\cite{ren2024xcube}, MeshFormer~\cite{liu2025meshformer}, LTS3D~\cite{meng2024lts3d} and Trellis~\cite{xiang2024trellis}, enables more efficient training for high-resolution shapes and scenes and better preservation of high-frequency geometric details. Meanwhile, OctFusion~\cite{xiong2024octfusion} proposes to represent a 3D shape as a volumetric octree with each leaf node encoded by latent features. Although these approaches offer adaptiveness in the latent representation similar to ours, their spatial structure is determined by volumetric occupancy rather than surface complexity.

\myparagraph{Octree-based 3D representation.} Octree~\cite{meagher1980octree,meagher1982geometric} is an efficient 3D data structure that recursively divides a 3D space into eight octants. It adapts to sparsity and minimizes storage and computation in empty regions, making it both memory- and computationally efficient. Compared to dense voxel grids, octrees significantly reduce memory usage while preserving fine geometric details in complex regions. %
Octree has been used in a wide range of geometric processing applications, including point cloud compression~\cite{schnabel2006octree}, 3D texturing~\cite{benson2002octree}, multi-view scene reconstruction~\cite{szeliski1993rapid,yu2021plenoctrees}, shape analysis~\cite{wang2017cnn,riegler2017octnet,Wang2023OctFormer}, and shape generation~\cite{tatarchenko2017octree,xiong2024octfusion,Ibing_2023_CVPR,wei2025octgpt}. 
While similar adaptive octree~\cite{wang2018adaptive} has been used for the shape classification and prediction tasks,
our work is the first to explore octree representation in the context of 3D tokenization and autoregressive generation, which requires us to co-design the encoding, decoding, and generation with octree data structure. 
Compared to existing approaches~\cite{ren2024xcube,xiong2024octfusion,xiang2024trellis} that use uniform tokenization schemes, our method adapts tokenization according to shape complexity. %

\begin{figure*}
    \centering
    \includegraphics[width=\linewidth]{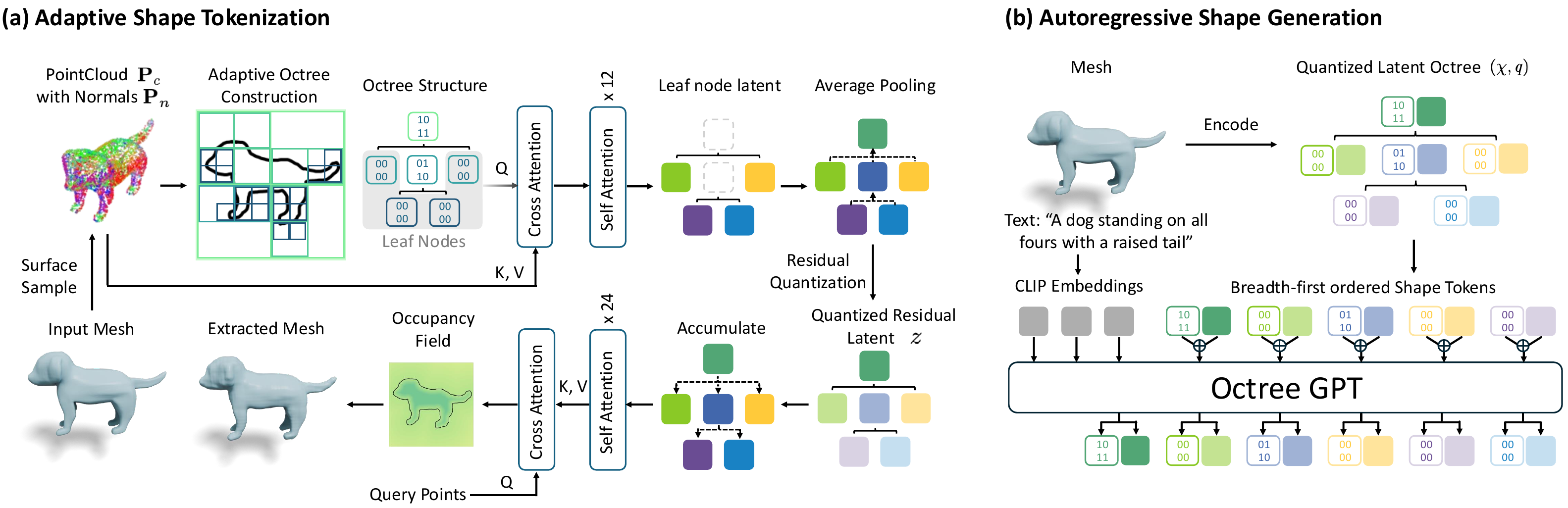}
    \vspace{-20 pt}
    \caption{%
    (a) {\bf Adaptive Shape Tokenization.}
   Given an input mesh with surface point samples, we partition 3D space into a sparse octree that adapts to the local geometric complexity of the surface. We then use a Perceiver-based transformer~\cite{jaegle2021perceiver}
 to encode the shape into a tree of latent codes, where a child node need encode only the (quantized) residual latent relative to its parent~\cite{lee2022residual}. Latents can then be decoded into an occupancy field from which a mesh can be extracted. (b) {\bf Autoregressive Shape Generation.} We define an autoregressive model for generating a tree of quantized shape tokens given a textual prompt, following a coarse-to-fine breadth-first search traversal. Similar to variable-length generation of text via end-of-sentence tokens, we make use of structural tokens to generate variable-size tree structures.
    }
    \vspace{-10 pt}
    \lblfig{overview}
\end{figure*}

\section{Method}
\lblsec{method}

\reffig{overview} illustrates our text-based shape generation framework. Our approach comprises two components: (1) a shape tokenization method (Octree-based Adaptive Tokenization, \ours) in \refsec{octree_construction} and \ref{sec:octoken} that efficiently compresses 3D shapes into a compact latent space, and (2) an autoregressive generative backbone model, \ourAR in \refsec{octreegpt}, which operates on these variable-length shape tokens.

For each 3D shape, our approach begins by sampling a point cloud $\rmP_c \in \mathbb{R}^{N \times 3}$ from the surface, along with its surface normal vectors $\rmP_n \in \mathbb{R}^{N \times 3}$ following prior work~\cite{zhang20233dshape2vecset}. We then employ our novel adaptive octree construction algorithm that partitions 3D space based on \emph{local geometric complexity} to obtain an sparse octree structure. We then leverage the Perceiver-based transformer architecture~\cite{jaegle2021perceiver} to encode the shape into an adaptive latent tree structure. The resulting variable-length latent representation can then be decoded into an occupancy field, from which a mesh can be extracted using marching cubes~\cite{lorensen1998marching}.

Unlike existing shape VAEs~\cite{zhang20233dshape2vecset,li2024craftsman,zhang2024clay,zhao2025hunyuan3d} that learn fixed-size latent representations for every shape using Variational Autoencoders~\cite{kingma2013vae}, we propose to encode shapes into variable-length latents based on their shape complexity. This adaptive tokenization approach results in a more compact latent space that only uses more latents by subdividing cells to finer resolution where the complexity of the shape is higher --- thereby leading to better reconstruction quality and improved performance in downstream generative tasks.

\subsection{Complexity-Driven Octree Construction}
\lblsec{octree_construction}

One of the core ingredients of our method is a sparse octree data structure which subdivides octants according to local geometry complexity, unlike existing methods subdividing cells based on occupancy. 

An octree is a hierarchical spatial data structure that recursively subdivides 3D space into eight equal octants. Starting with a root node representing a bounding cube, each non-empty node can be further partitioned into eight child nodes, creating a tree-like structure $\gO = \{\gV, \gE\}$. We use $\gV = \{v_1, v_2, \dots\}$ to denote the cells in an octree hierarchy, and $\gE \subseteq \gV \times \gV$ defines parent-child relationships, where $(v_i, v_j) \in \gE$ indicates that $v_i$ is the parent of $v_j$. This representation is efficient to represent sparse 3D data, as it allocates higher resolution only to occupied regions. %

In this paper, we consider the \emph{sparse} octree by omitting empty child nodes, i.e., each node can have 0 to 8 child nodes, with all nodes being non-empty. This structure can be compactly encoded by an 8-bit binary code $\chi: \gV \rightarrow \{0,1\}^8$. For instance, $\chi(v) = \left(01001000\right)_2$ indicates that node $v$ has two non-empty child nodes at its second and fifth slots. An octree structure can thus be {\em uniquely} represented as a sequence of 8-bit binary codes in breadth-first order, $\left[\chi(v_0), \chi(v_1), \cdots\right]$.

While octrees have previously been used to tokenize 3D shapes, earlier methods~\cite{xiong2024octfusion, ren2024xcube} always subdivide up to the maximum depth unless empty. 
In contrast, we subdivide an octant only when the local geometry is ``complex''. Inspired by the literature in mesh simplification \cite{GarlandH97} and isosurfacing \cite{JuLSW02}, we use the quadric error metric to measure shape complexity and guide octree subdivision. This approach optimizes representational capacity, allocating tokens where they provide the greatest benefit for shape fidelity.

\myparagraph{Quadric error metric} was first introduced to quantify local geometric complexity for mesh simplification tasks~\cite{garland2005quadric}. Given a plane in $\mathbb{R}^3$, let $\rvp$ denote a point on the plane with unit normal vector $\rvn$. The plane can be defined by all points $\rvx \in \mathbb{R}^3$ satisfying
\begin{equation}
    \rvn^\top (\rvx - \rvp) = 0.
\end{equation}

The quadric error measures the squared point-to-plane distance between
any point $\rvx$ and this plane, computed as
\begin{align}
    E(\rvx) = \left( \rvn^\top (\rvx - \rvp) \right)^2 \doteq [\rvx^\top,1] \:  \rmQ \: [\rvx^\top,1]^\top,
\end{align}
where the quadric matrix $\rmQ \in \mathbb{R}^{4\times 4}$ is defined as
\begin{equation}
    \rmQ = \left[ 
    \begin{matrix}
    \rvn\rvn^\top & - \rvn\rvn^\top\rvp \\
    (-\rvn\rvn^\top\rvp)^\top & \rvp^\top\rvn\rvn^\top\rvp 
    \end{matrix}
    \right].
\end{equation}

As a key property, the cumulative error from a point $\rvx$ to {\em multiple} planes can be computed with a summed quadric,
\begin{equation}
    E(\rvx) = \sum_i E_i(\rvx) = [\rvx^\top,1] \:  \left(\sum_i\rmQ_i\right) \: [\rvx^\top,1]^\top.
\end{equation}

We use the quadric error $E^* = \min_{\rvx } E(\rvx)$ to measure local geometric complexity. As the energy is quadratic, the minimum $E^*$ can be efficiently computed by solving a linear system, with details left in the appendix. Intuitively, when the planes form common intersections (e.g., an edge, a cone, or being flat), the optimal quadric error approaches zero, whereas complex regions usually yield higher quadric error values. This property makes quadric error metrics suitable for guiding adaptive geometric representations.

Specifically, for each octree cell $v \in \gV$, we compute the cell quadric $\rmQ_v$ by summing up quadrics for all sampled points within $v$,
\begin{equation}
    \rmQ_v = \sum_{\rvp_ \in \rmP_c(v)} \rmQ_{\rvp},
\end{equation}
where $\rmP_c(v) = \{ \rvp \in \rmP_c \:\vert\: \rvp \text{ is contained in cell } v\}$ denotes the subset of points that lie within cell $v$, and 
$\rmQ_p$ is the quadric matrix for point $\rvp$ with its corresponding normal vector $\rvn \in \rmP_n$. We then calculate the average quadric error
\begin{equation}
    E_v^* = \min_{\rvx} E_v(\rvx) = \frac{1}{|\rmP_c(v)|} \min_{\rvx } [\rvx^\top,1]  \rmQ_v [\rvx^\top,1]^\top.
\end{equation}
We recursively subdivide $v$ into child cells only when both of these conditions are met: (1) the maximum depth $L$ has not been reached, and (2) the quadric error exceeds our pre-defined threshold, $E_v^* > T$. %
In regions with complex geometry, cells are subdivided to the maximum depth $L$, while subdivision stops early in areas with simpler (i.e., planar) geometry. %

\subsection{Adaptive shape tokenization with \ours}
\lblsec{octoken}

Following prior work~\cite{zhang20233dshape2vecset,zhao2023michelangelo,li2024craftsman,zhang2024clay,zhao2025hunyuan3d}, we adopt a Perceiver-based variational autoencoder (VAE)~\cite{jaegle2021perceiver,kingma2013vae} to encode the shape into latents. Specifically, we compute:  
\begin{align}
    \hat\rmP & = \text{Concat}\left(\text{PE}(\rmP_c), \rmP_n\right), \\
    \hat\rmO & = \text{Concat}\left(\text{PE}(\gV_\text{leaf}), \text{SE}(\gV_\text{leaf})\right), \\
    \phi(\gV_\text{leaf}) & = \text{SelfAttn}^{(i)}(\text{CrossAttn}(\hat\rmO, \hat\rmP)), \,  i = 1, \cdots, L_e, \nonumber
\end{align}
where the encoder $\phi$ outputs a latent vector $\phi(v)$ for every leaf cell $v \in \gV_\text{leaf}$,  where $\phi: \gV \rightarrow \mathbb{R}^d$. 
Here, PE denotes the positional encoding function~\cite{vaswani2017attention}, which operates on point coordinates and octree cell centers, while SE denotes the scale encoding function on the depth of octree cells.  $\gV_\text{leaf}$ comprises all the leaf cells within $\gV$, and $L_e$ refers to the number of Self Attention layers in the shape encoder.

Notably, the cross-attention operation is global, allowing each leaf cell to attend to all points in $\hat\rmP$ across the entire shape,  rather than just points within its local cell. This global attention enables the model to capture long-range dependencies and contextual information beyond local neighborhoods. The subsequent self-attention layers further refine these representations by allowing leaf cells to exchange information. 

Finally, we propagate latent vectors from leaf cells to their ancestors bottom-up. Each non-leaf node computes its latent vector by averaging those of its child nodes.

\myparagraph{Multi-scale octree residual quantization.} The variable length of the encoded latent motivates us to adopt an autoregressive model for downstream generation in \refsec{octreegpt}. This approach requires us to learn a quantization bottleneck in the VAE.
To achieve this, %
we propose an octree-based residual quantization strategy, enabling a coarse-to-fine token ordering using residual quantization~\cite{tian2025var,lee2022residual}. Specifically, we start quantization from the root node and only process the residual latent of every latent from its parent. We use a shared codebook and quantization function for all of the nodes using vqtorch~\cite{huh2023vqtorch}.
We summarize our residual quantization algorithm in \refalg{encode}.

\begin{algorithm}[t]
{
\small
\caption{Multi-scale octree residual quantization}\lblalg{encode}
\begin{algorithmic}[1]
\Require Octree $\gO=\{\gV,\gE\}$, Latent $\phi: \gV \rightarrow \mathbb{R}^d$.
\Ensure Multi-scale residual quantized latent $z: \gV \rightarrow \mathbb{R}^d$, Quantized latent index $q: \gV \rightarrow \mathbb{Z}$.
\State $z(v_0), q(v_0) = \text{Quantize}(\phi(v_0))$ \Comment{$v_0$ is the root node.}
\State $z_{acc}(v_0) = z(v_0)$ \Comment{Initialize accumulated latent.}
\For{$d = 1, \cdots, L-1$} \Comment{L is the max depth of $\gO$.}
\For{$v \in \gV_d$} \Comment{$\gV_d$ is the set of nodes at level $d$.}
\State Find the parent $v_{\text{parent}}$ of $v$ according to $\gE$.
\State $z(v), q(v) = \text{Quantize}(\phi(v) - z_{acc}(v_\text{parent}))$.
\State $z_{acc}(v) = z_{acc}(v_\text{parent}) + z(v)$. \Comment{Update $z_{acc}$.}
\EndFor 
\EndFor
\end{algorithmic}
}
\end{algorithm}

\myparagraph{Octree decoding.} Given the multi-scale octree residual latent $z: \gV \rightarrow \mathbb{R}^d$, we recover the full latent $\hat\phi: \gV \rightarrow \mathbb{R}^d$ by adding the latent to every node from all its ancestors. Motivated by prior work~\cite{zhang20233dshape2vecset,li2024craftsman,zhao2023michelangelo}, we use a similar perceiver-based transformer to decode the latent to an occupancy field. Specifically, given a query 3D point $\rvx \in \mathbb{R}^3$, the decoder predicts its occupancy value:
\begin{align}
    \hat\rmS & = \text{Concat}\left(\hat\phi(\gV), \text{PE}(\gV), \text{SE}(\gV)\right), \\
    \hat\rmS & = \text{SelfAttn}^{(j)}(\hat\rmS), \qquad j = 1, 2, \cdots, L_d, \\
     \hat\sigma(\rvx, \hat\phi, \gO) & = \text{CrossAttn}\left(\text{PE}(\rvx), \hat\rmS\right),
\end{align}
where $L_d$ is the number of Self Attention layers in the shape decoder, and $\hat\sigma$ is the predicted occupancy value at the query point. At inference time, we query the decoder using grid points and run marching cubes~\cite{lorensen1998marching} to extract a mesh. During training, we sample query points using uniform and importance sampling near the mesh surface following prior work~\cite{zhang20233dshape2vecset,zhao2023michelangelo,li2024craftsman}.%
We jointly optimize the networks and codebook via the following loss functions. %
\begin{align}
\mathcal{L}_{\text{VQ}} & = \mathbb{E}_{v \in} \vert\vert \text{sg}(\hat\phi(v)) - \phi(v) \vert\vert^2 + \vert\vert \text{sg}(\phi(v)) - \hat\phi(v) \vert\vert^2, 
\end{align}
where $\text{sg}()$ is the stop-gradient operation.  Additionally, we incorporate an occupancy reconstruction loss to ensure that the latent codes accurately reconstruct the input shape:  
\begin{align}
\mathcal{L}_{\text{rec}}  = \mathbb{E}_{\rvx} \mathcal{L}_{\text{BCE}}\left(\sigma(\rvx), \hat\sigma(\rvx, \hat\phi, \gO)\right), 
\end{align}
where $\mathcal{L}_{\text{BCE}}$ is the binary cross-entropy loss for shape reconstruction, and $\sigma(\rvx) \in \{0,1\}$ is the ground truth occupancy value of the query point, indicating whether it is located inside the object.
Our final loss function is
$\mathcal{L}_{\text{rec}}
 + \lambda_{\text{VQ}} \mathcal{L}_{\text{VQ}}$,
where $\lambda_{\text{VQ}}$ weights the vector quantization loss.

\myparagraph{KL variant for continuous tokens.} By replacing the quantization bottleneck with a KL regularization~\cite{kingma2013vae},
our proposed \ours can learn continuous shape latent instead, which provides a fair comparison with other continuous latent baselines in \refsec{exp_recon}.

\subsection{OctreeGPT: Autoregressive Shape Generation}
\lblsec{octreegpt}

Building on our adaptive tokenization framework, we develop OctreeGPT, an autoregressive model for generating 3D shapes conditioned on text descriptions. Unlike previous approaches that operate on fixed-size representations~\cite{li2024craftsman,zhang2024clay,zhao2025hunyuan3d}, OctreeGPT models the joint distribution of variable-length octree tokens while maintaining a hierarchical coarse-to-fine structure.

\myparagraph{Shape Token Sequence.} To enable autoregressive modeling, we first serialize the octree structure by traversing it in a breadth-first manner as mentioned in \refsec{octoken}. For each node $v$, we include both its quantized index $q(v) \in \mathbb{Z}$ and a structural code $\chi(v) \in \{0,1\}^8$ that encodes the presence or absence of each potential child node. A latent octree can thus be {\em uniquely} represented by a variable-length sequence of tokens:
    $\left[t_0, t_1, \cdots, t_N \right],$
where each token $t_i = \left(q(v_i), \chi(v_i)\right), \forall i \in \mathbb{N}$.

We train an autoregressive model that predicts the next token in the sequence,
\begin{equation}
    P\left(t_0, t_1, \cdots, t_N | \theta\right) = \prod_{i=1}^{N} P(t_i | t_0, \cdots, t_{i-1}, \theta),
\end{equation}
where $\theta$ is our learned OctreeGPT model.

\myparagraph{Model Architecture.} Our architecture builds upon decoder-only transformers similar to GPT-2~\cite{esser2021taming,radford2019gpt}. Specifically, we compute the embedding for each shape token $t_i$ as:
\begin{equation}
    \text{Embed}(t_i) = \text{Embed}_{q}(q(v_i)) + \text{Embed}_{\chi}(\chi(v_i)) + \text{PE}_{\text{tree}}(v_i),
\end{equation}
where $\chi(v_i)$ is interpreted as an 8-bit integer. The tree-structured positional encoding $\text{PE}_{\text{tree}}(v_i)$ captures both spatial and hierarchical information:
\begin{align}
    \text{PE}_{\text{tree}}(v_i) = \text{Embed}_{x}(x(v_i)) + \text{Embed}_{y}(y(v_i)) \\
    + \text{Embed}_{z}(z(v_i)) + \text{Embed}_{d}(d(v_i)), 
\end{align}
where $x,y,z$ are quantized coordinates of the cell center, and $d \in \{0,1,\cdots,L-1\}$ is the depth of the octree node. This multi-dimensional positional encoding helps the model understand both spatial relationships and the hierarchical structure of the octree.
Our model employs dual prediction heads for predicting quantized latent indices $\hat q$ and structural codes $\hat\chi$, allowing the model to jointly reason about geometry and tree structure.
For text-conditioned generation, we prepend the sequence with 77 tokens derived from the input text's CLIP embedding~\cite{radford2021clip}.

\myparagraph{Training and Inference.} We train OctreeGPT using a combined loss function that balances the reconstruction of latent tokens and structural codes:
\begin{equation}
\mathcal{L}_{\text{GPT}} = \mathcal{L}_{\text{CE}}(q, \hat{q}) + \lambda_{\chi} \mathcal{L}_{\text{CE}}(\chi, \hat{\chi}),
\end{equation}
where $\mathcal{L}_{\text{CE}}$ is the cross-entropy loss for $2^8$-way classification, and $ \lambda{\chi}$ are balancing hyperparameters.
During inference, we employ sampling with temperature $\tau$ to control the diversity and quality of generated shapes. We process the predicted structural code $\chi(v_i)$ on the fly to determine the octree topology, which dynamically establishes the final length of the token sequence.
Further implementation details and hyperparameter settings are provided in the appendix.

\begin{figure}[t]
    \centering
    \includegraphics[width=\linewidth]{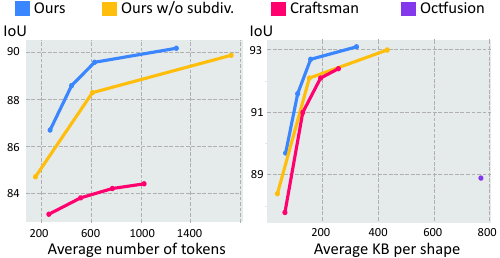}
    \vspace{-20 pt}
    \caption{We plot reconstruction quality (IoU) against latent size in both discrete (left) and continuous (right) scenarios. We use KiloBytes (KB) for continuous latent representations for a fair comparison.
    Our method consistently outperforms baseline approaches at equivalent latent sizes and achieves comparable reconstruction quality with much smaller latent representations.}
    \vspace{-10 pt}
    \lblfig{plot}
\end{figure}

\begin{figure}[t]
    \centering
    \includegraphics[width=\linewidth]{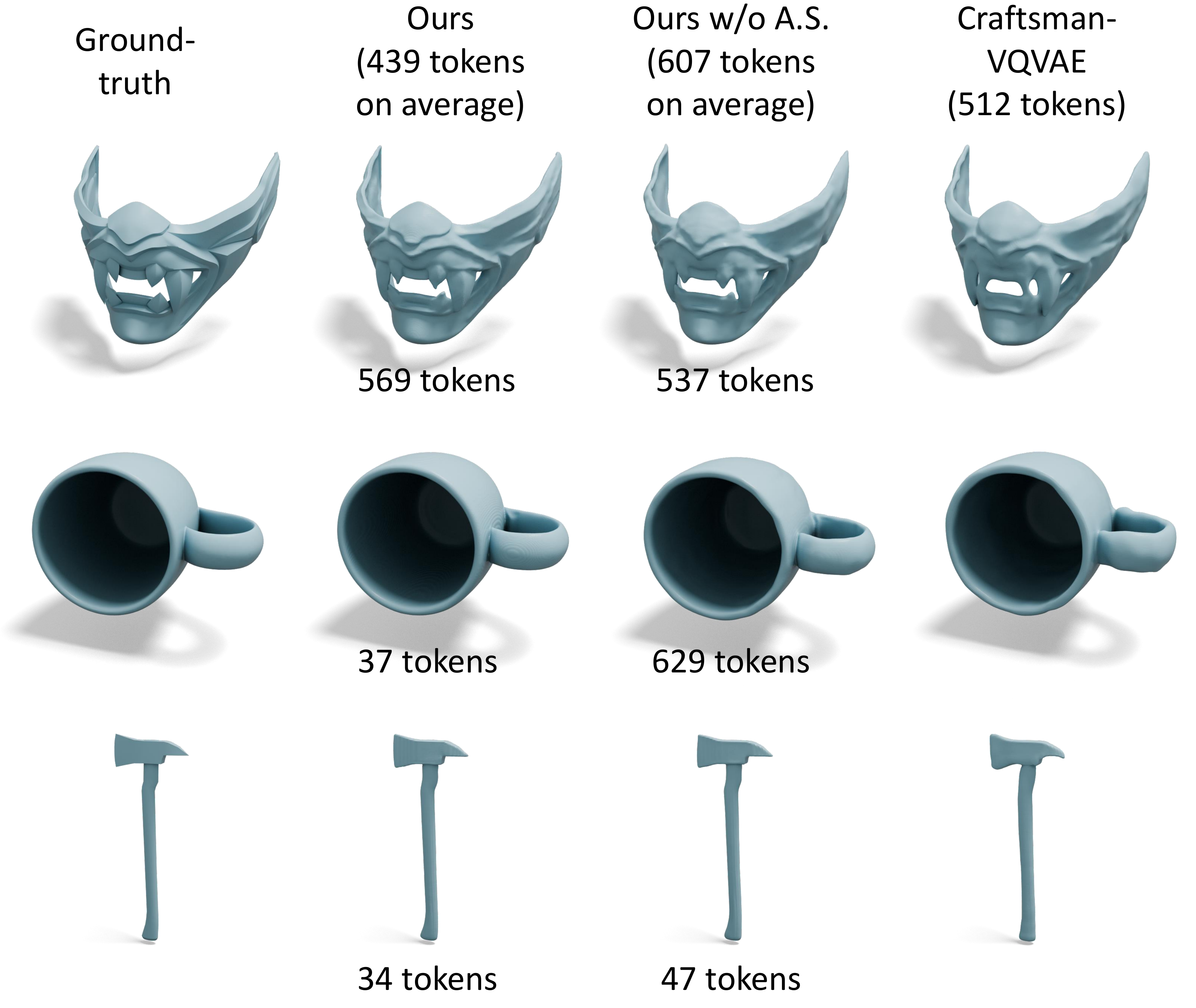}
    \vspace{-20 pt}
    \caption{\textbf{Shape reconstruction with discrete latent.} We compare our full method against Craftsman-VQ~\cite{li2024craftsman} as well as an ablation without Adaptive Subdivision (A.S.). With comparable or lower token budget, our method generally outperforms the baseline regarding reconstruction fidelity. Meanwhile, without adaptive subdivision, the vanilla octree only allocates the token budget efficiently for objects of small volume (bottom) but wastes tokens on geometrically simple objects that occupy large space (middle).}
    \vspace{-15 pt}
    \lblfig{comparison_vq}
\end{figure}

\begin{figure}[t]
    \centering
    \includegraphics[width=\linewidth]{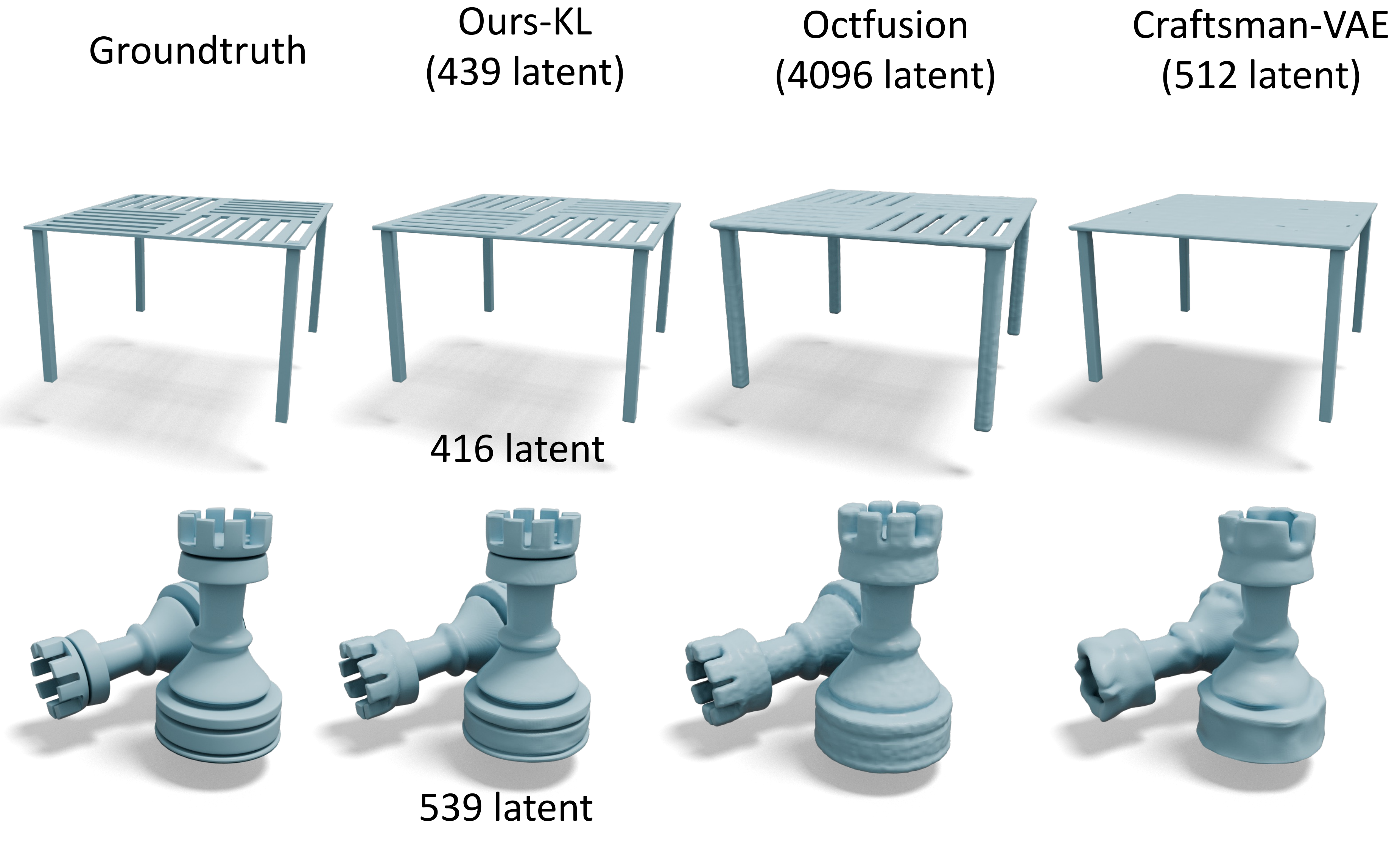}
    \vspace{-20 pt}
    \caption{\textbf{Shape reconstruction with continuous latent.}  
    We include the visual comparison between our continuous VAE (\ours-KL) and other baselines. In general, our reconstruction preserves more details using similar or smaller number of latent vectors.
    }
    \vspace{-10 pt}
    \lblfig{comparison_kl}
\end{figure}

\begin{figure}[t]
    \centering
    \includegraphics[width=\linewidth]{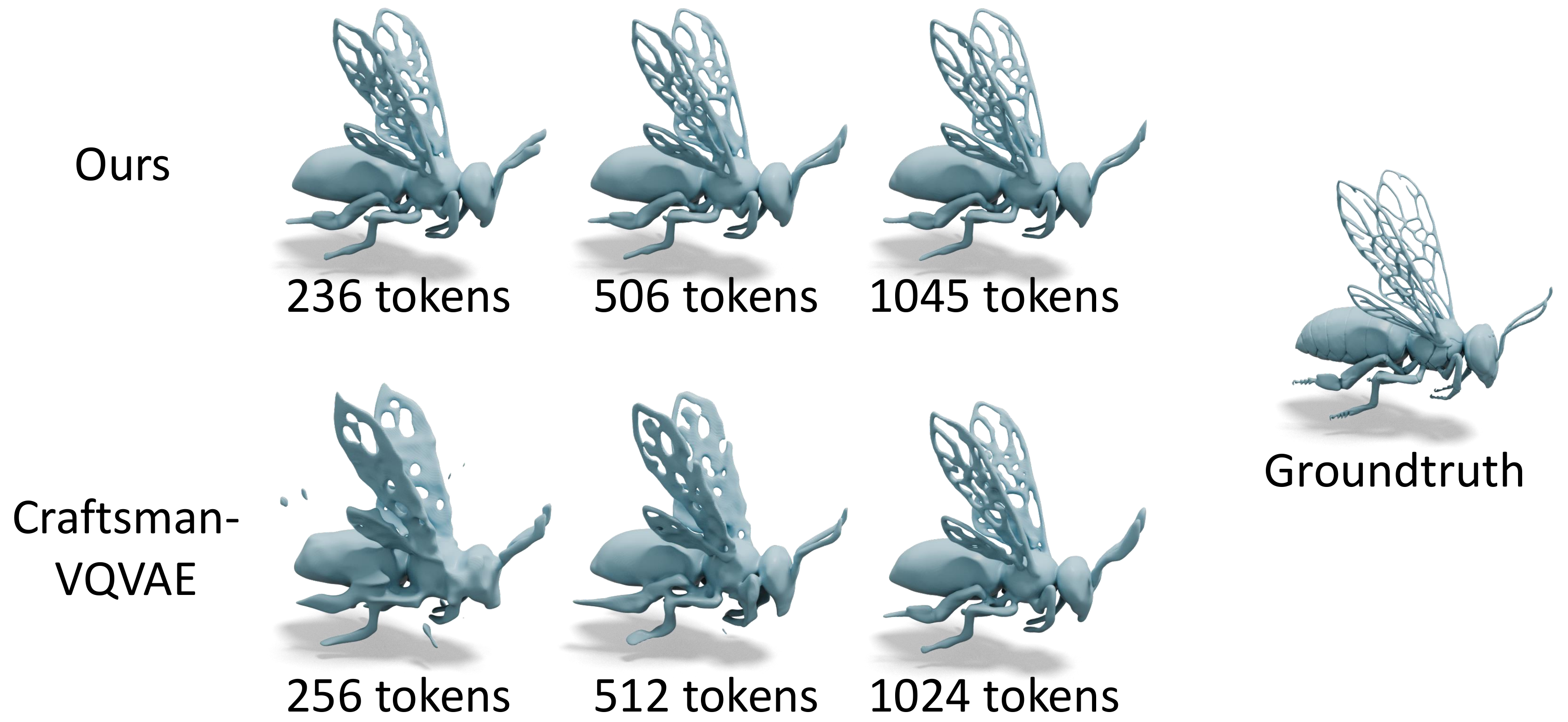}
    \vspace{-15 pt}
    \caption{\textbf{Ablation study on token length.} With an increasing number of tokens, our method achieves better quality while consistently outperforming the baseline at a comparable token length.}
    \vspace{-10 pt}
    \lblfig{abl_token}
\end{figure}

\begin{table}[t]
\setlength{\tabcolsep}{4pt}
\small
  \begin{tabular}{lccc}
    \toprule
    Method & Avg Token Cnt & IoU $\uparrow$ & CD ($\times 10^{-3}$) $\downarrow$ \\
    \midrule
      \multirow{4}{*}{Craftsman-VQ~\cite{li2024craftsman}}  & 256 & 83.1 & 2.31 \\
     & 512 & 83.8 & 1.94 \\
     & 768 & 84.2 & 1.88 \\
     & 1024 & 84.4 & 1.80 \\
    \midrule
    \multirow{3}{*}{\shortstack[l]{\textbf{Ours (\ours)} \\ w/o A.S.}} & 148 & 84.7 & 2.19\\
    & 607 & 88.3 & 1.85 \\
    & 1726 & 89.9 & 1.37\\
    \midrule
     \multirow{4}{*}{\textbf{Ours (\ours)}} & 266 & 86.7 & 1.94\\
     & 439 & 88.6 & 1.78 \\
     & 625 & 89.7 & 1.53 \\
     & 1284 & \textbf{90.2} & \textbf{1.27} \\

  \bottomrule
\end{tabular}
\vspace{-5 pt}
\caption{\textbf{Quantitative analysis of shape reconstruction with discrete latent.} We compare our method against Craftsman-VQ~\cite{li2024craftsman} and ablation without Adaptive Subdivision (A.S.). With comparable token counts, our approach outperforms both baselines, showing the effectiveness of our proposed adaptive tokenization.}
\lbltbl{recon_vq}
\end{table}

\begin{table}[t]
\setlength{\tabcolsep}{4pt}
\small
  \begin{tabular}{lccc}
    \toprule
    Method & Avg Latent Len & IoU $\uparrow$ & CD ($\times 10^{-3}$) $\downarrow$ \\
    \midrule
      \multirow{4}{*}{Craftsman~\cite{li2024craftsman}}  & 256 & 87.8 & 1.96 \\
     & 512 & 91.0 & 1.83 \\
     & 768 & 92.1 & 1.33 \\
     & 1024 & 92.4 & 1.29 \\
    \midrule
    Octfusion$^\dagger$~\cite{xiong2024octfusion} & 4096 & 88.9 & 1.87 \\
    XCube$^\dagger$~\cite{ren2024xcube} & 4096 & - & 1.26 \\
    \midrule
    \multirow{3}{*}{\shortstack[l]{\textbf{Ours (\ours-KL)} \\ w/o A.S.}} & 148 & 88.4 & 1.89\\
    & 607 & 92.1 & 1.29 \\
    & 1726 & 93.0 & 1.01 \\
    \midrule
     \multirow{4}{*}{\textbf{Ours (\ours-KL)}} & 266 & 89.7 & 1.81\\
     & 439 & 91.6 & 1.29\\
     & 625 & 92.7 & 1.08\\
     & 1284 & \textbf{93.1} & \textbf{0.97}\\
  \bottomrule
\end{tabular}
\vspace{-5 pt}
\caption{\textbf{Quantitative analysis of shape reconstruction with continuous latent.} We replace the quantization with a KL regularization to learn continuous latent (\ours-KL) as mentioned in \refsec{octoken}. Our method outperforms all the baselines with comparable or shorter latent code lengths.
$\dagger$ indicates off-the-shelf models that are pre-trained on different data sources than ours.}
\vspace{-10pt}
\lbltbl{recon_kl}
\end{table}

\begin{table}[t]
\setlength{\tabcolsep}{4pt}
\small
  \begin{tabular}{lcccc}
    \toprule
    \multirow{2}{*}{Method} & \multirow{2}{*}{FID$\downarrow$} & KID$\downarrow$  & CLIP- &  Runtime$\downarrow$  \\ %
    & & ($\times10^{-3}$) & score$\uparrow$ & (secs) \\
    \midrule
    Craftsman$^\dagger$~\cite{li2024craftsman} & 65.18 & 6.42 & 0.27 & 54.8 \\
    InstantMesh$^\dagger$~\cite{mueller2022instant} & 67.93 & 7.23 & 0.31 & 21.5  \\
    XCube~\cite{ren2024xcube} & 132.56 & 9.83 & 0.23 & 32.3 \\
    Craftsman-VQ + GPT & 85.10 & 7.49 & 0.26 & 15.4 \\
    \midrule
    Ours (\ourAR)  & \textbf{56.88} & \textbf{5.79} & \textbf{0.34} & \textbf{11.3}\\
  \bottomrule
\end{tabular}
\vspace{-5 pt}
\caption{\textbf{Quantitative analysis of shape generation.} We compare \ourAR with a GPT baseline trained on Craftsman-VQ (\refsec{exp_recon}), XCube~\cite{ren2024xcube}, and image-to-3D methods InstantMesh~\cite{xu2024instantmesh} and Craftsman~\cite{li2024craftsman}. We compute FID~\cite{heusel2017fid}, KID~\cite{binkowski2018kid}, and CLIP-score on the renderings of generated shapes, and report the average generation time. Our method outperforms all the baselines, showing higher quality and better consistency with the input text while achieving the fastest runtime due to our efficient tokenization.
}
\vspace{-15 pt}
\lbltbl{gen}
\end{table}

\begin{figure*}[t]
    \centering
    \includegraphics[width=\linewidth]{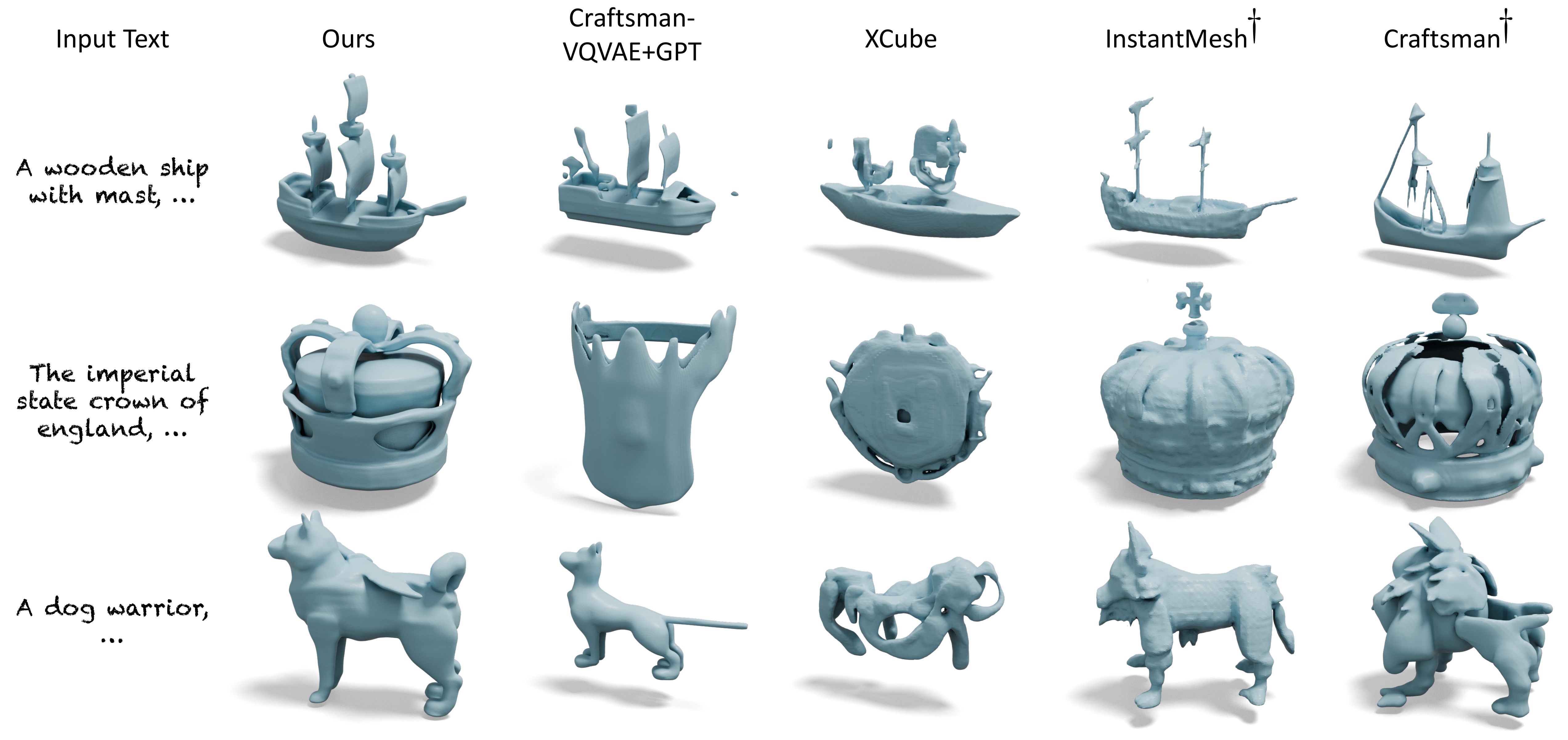}
    \vspace{-20 pt}
    \caption{\textbf{Shape Generation Results.} We compare OctreeGPT with a GPT baseline trained on Craftsman-VQVAE (\refsec{exp_recon}), text-to-3D model XCube~\cite{ren2024xcube}, and image-to-3D methods InstantMesh~\cite{xu2024instantmesh} and Craftsman~\cite{li2024craftsman}. Our results have smoother surfaces, finer details, and fewer artifacts than baselines. For image-conditioned methods$^\dagger$, we use FLUX.1~\cite{fluxschnell} to generate condition images from input text.
    }
    \vspace{-10 pt}
    \lblfig{gen}
\end{figure*}

\section{Experiments}
\lblsec{experiment}

We evaluate our method on shape tokenization and generation. We perform qualitative and quantitative comparisons with existing baselines and conduct an ablation study on the significance of each major component.

\myparagraph{Dataset.} 
We use the Objaverse~\cite{objaverse} dataset, which contains around 800K 3D models, as our training and test data. To ensure high-quality training and evaluation, we filter out low-quality meshes, such as those with point clouds, thin structures, or holes. This results in a curated dataset of around 207K objects for training and 22K objects for testing. 

For preprocessing, each mesh is normalized to a unit cube.  
For each mesh, we sample 1M points with their normals from the surface as the input point cloud. To generate ground-truth occupancy values, we uniformly sample 500K points within the unit volume and an additional 1M points near the mesh surface to capture fine details and obtain the occupancy based on visibility following prior work~\cite{roblox2025cube}. 
We then construct an adaptive octree for each shape based on the sampled point cloud using a pre-defined quadric error threshold $T$, which guides the subdivision process according to local geometric complexity.
To enable text conditioning, we render nine views of each object under random rotations and use GPT-4o~\cite{achiam2023gpt} to generate descriptive captions from these renderings. 

\subsection{Shape Reconstruction}
\lblsec{exp_recon}

\myparagraph{Baselines.} We compare \ours with latent vector sets from Craftsman3D~\cite{li2024craftsman}.
For a fair comparison, we train both methods under identical conditions,  using both quantization for discrete tokenization and KL regularization for continuous latent space.
Additionally, we evaluate against two other recent approaches, XCube~\cite{ren2024xcube} and Octfusion~\cite{xiong2024octfusion}.
Due to computational resource constraints, we use publicly available pre-trained models for these two baselines rather than retraining them on our dataset.
We exclude VAE models from Direct3D~\cite{wu2024direct3d}, CLAY~\cite{zhang2024clay}, and LTS3D~\cite{meng2024lts3d} as their implementations are not available.

\myparagraph{Results.} We evaluate shape reconstruction quality using volume Intersection over Union (IoU) and Chamfer Distance (CD) with 10K sampled surface points in \reftbl{recon_vq} and \reftbl{recon_kl}. Note that XCube~\cite{ren2024xcube} outputs an Unsigned Distance Function (UDF), which cannot be evaluated with IoU metrics. Visual comparisons in \reffig{comparison_vq} and \reffig{comparison_kl} demonstrate our approach outperforms all baselines.

\myparagraph{Ablation Study.} We ablate our proposed adaptive subdivision in \reffig{comparison_vq}. Without quadric-error-based adaptive subdivision, the octree representation subdivides to the deepest level unless empty, wasting tokens on simple objects of large volumetric occupancy (middle row). \reffig{plot} shows reconstruction quality (IoU) versus latent size in both discrete and continuous scenarios, confirming our method achieves better quality at equivalent latent sizes and requires significantly smaller latent representations for comparable reconstruction quality. \reffig{abl_token} further shows a qualitative comparison between our method and the baseline in reconstruction quality with respect to the number of tokens used.

\subsection{Shape Generation}
\lblsec{exp_gen}

This section evaluates our text-to-shape generation quality against multiple baselines. We train our \ourAR on top of \ours using 439 tokens on average, and for comparison, train a GPT model on Craftsman-VQ with 512 tokens. We include XCube~\cite{ren2024xcube}'s pre-trained Objaverse model as a native text-to-3D baseline. We also compare against two image-to-3D methods, InstantMesh~\cite{xu2024instantmesh} and Craftsman~\cite{li2024craftsman}, using FLUX.1~\cite{fluxschnell} to generate condition images from input text.

\myparagraph{Results.} We quantitatively evaluate generation quality in \reftbl{gen} by rendering generated shapes and computing FID~\cite{heusel2017fid,parmar2021cleanfid} and KID~\cite{binkowski2018kid} against groundtruth renderings. We also report CLIP-score~\cite{radford2021clip} to evaluate text-shape consistency, and average generation time to evaluate efficiency. In addition, we also provide qualitative comparisons in \reffig{gen}. Overall, thanks to a more compact and representative latent space, our \ourAR produces finer details with fewer artifacts compared to Craftsman-VQ with GPT, while also outperforming other 3D generation baselines in both geometry quality and prompt adherence, with a faster runtime.

\section{Discussion}

In this work,  we propose an octree-based Adaptive Shape Tokenization, \ours,  a framework that dynamically adjusts latent representations according to shape complexity. At its core, \ours constructs an adaptive octree structure guided by a quadric-error-based subdivision criterion, allocating more tokens to complicated parts and objects while saving on simpler ones. Extensive experiments show that \ours reduces token counts by 50\% compared to previous fixed-size approaches while maintaining comparable visual quality. Alternatively, with a similar number of tokens, \ours produces much higher-quality shapes. Building upon this tokenization, we develop an octree-based Autoregressive generative model, \ourAR that effectively leverages these variable-sized representations, outperforming existing baselines.

\myparagraph{Limitations.} 
Our framework only addresses geometric shape reconstruction and generation without incorporating texture information. We leave modeling both shape and texture properties jointly for future work.

\clearpage

\section*{Acknowledgement}
We thank Akash Garg, Daiqing Li, Alexander Weiss, Alejandro Peláez, Kayvon Fatahalian, Sheng-Yu Wang, Gaurav Parmer, Ruihan Gao, Nupur Kumari, and Maxwell Jones for their discussion and help. This work was done when KD was an intern at Roblox. The project is partly supported by Roblox. JYZ is partly supported by the Packard Fellowship. The Microsoft Research PhD Fellowship supports KD. 

{
    \small
    \bibliographystyle{ieeenat_fullname}
    \bibliography{main}
}

\clearpage

\appendix
\section*{Appendix}
\section{Quadric Error Computation}
In Section 3.1, we employ quadric error to guide our adaptive octree subdivision.
Quadric error is, by definition, the minimizer of a quadratic energy 
\begin{equation}
    E(\rvx) = [\rvx^\top,1] \:  \rmQ \: [\rvx^\top,1]^\top,
\end{equation}
characterized by the quadric matrix $\rmQ$
\begin{equation}
    \rmQ = \left[ 
    \begin{matrix}
    \rvn\rvn^\top & - \rvn\rvn^\top\rvp \\
    (-\rvn\rvn^\top\rvp)^\top & \rvp^\top\rvn\rvn^\top\rvp 
    \end{matrix}
    \right] = \left[ 
    \begin{matrix}
    \rmA  & \rvb \\
    \rvb^\top & c 
    \end{matrix}
    \right],
\end{equation}
where we follow the same notation as in the main text, using $\rvp,\rvn \in \mathbb{R}^3$ to denote the location and the normal vector, respectively.
For clarify purposes, we use $\rmA, \rvb, c$ to abbreviate the expression.

As the energy $E(\rvx)$ is quadratic, one can compute the minimizer $\rvx^\star$ by setting the derivative to zero
\begin{equation}
    \frac{\partial E(\rvx)}{\partial \rvx} = 0,
\end{equation}
which amounts to solve a 3-by-3 linear system in the form of
\begin{equation}
    \rmA \rvx^\star = -\rvb
\end{equation}
If the matrix $\rmA$ is well-conditioned, one can solve
for $\rvx^\star$ with standard solvers, such as Cholesky decomposition. If not, numerical surgeries, such as singular value decomposition \cite{Lindstrom00} or Tikhonov regularization \cite{TrettnerK20}, are recommended to solve for $\rvx^\star$. 
Once we obtain the minimizer $\rvx^\star$, we can then compute the quadric error by evaluating $E(\rvx^\star)$ at the optimal location.

\section{Implementation Details}

\myparagraph{Hyperparameters.} We provide our choice of quadric error threshold $T$ and Octree max depth $L$ in \reftbl{hyper}.
For shape VAE learning, we use 768 as the width and a codebook size of 16384. We set the maximum token length to 2048 and trim those exceeding samples. Since our octree nodes are sorted based on breadth-first order, we will only omit leaf nodes which will not affect the overall octree structure.
We set both $\lambda_{\text{VQ}}$ and $\lambda_{\chi}$ to 1.0.
We train our \ours with a batch size of 24 on 8 GPUs for 300K iterations. We train our \ourAR with a batch size of 16.
For both experiments, we use AdamW optimizer with a learning rate of 3e-4. 

\section{Additional Results}

\begin{table}[h]
\setlength{\tabcolsep}{6pt}
\centering
\scriptsize
\begin{tabular}{l|ccc}
\toprule
  Method & Avg. Token & Runtime (secs) $\downarrow$ & Memory (GB) $\downarrow$ \\
  \midrule
Craftsman-VQ+GPT  & 512 & 15.4 {\color{gray}\scriptsize±0.3} & 25.6{\color{gray}\scriptsize±0.8} \\
Ours ($T=5\text{e-}4$)  & 439 & 11.3 {\color{gray}\scriptsize±2.7}  & 20.4{\color{gray}\scriptsize±5.3} \\
Ours ($T=1\text{e-}3$) & 266 & \textbf{7.9} {\color{gray}\scriptsize±1.4} & \textbf{13.6}{\color{gray}\scriptsize±3.8} \\
\bottomrule
\end{tabular}
\caption{\textbf{Memory usage and runtime analysis.}}
\lbltbl{memory}
\end{table}

\myparagraph{Memory usage and runtime analysis.} \reftbl{memory} shows our shape generation runs faster and requires less VRAM on an A100 GPU, where $T$ is the quadric error threshold. We also profile the computational overhead of constructing an adaptive octree. Compared to a non-adaptive octree (0.12{\color{gray}\scriptsize±0.08} secs) that subdivides each octant containing any mesh element, our adaptive octree construction actually runs faster (0.09{\color{gray}\scriptsize±0.07} secs) on 100K points when $T=5\text{e-}4$. This efficiency comes from pruning low-complexity nodes and the rapid computation of quadric error as detailed in Sec. A.

\myparagraph{Ablation study on octree node ordering.} In Section 3.3, we train our \ourAR on the sequence of shape tokens from a latent octree. In our experiments, we empirically find that breadth-first ordering works the best. We present an ablation study in \reftbl{abl_order}. We find depth-first ordering works worse than breadth-first ordering potentially because of the lack of a coarse-to-fine scheme. We also experimented with the next scale prediction method from VAR~\cite{tian2025var}, where we simultaneously predict the tokens at the next level. While this method runs significantly faster, it produces much worse results than ours. We suspect it is due to the larger discrepancy ($2\times$) between levels in the octree structure while VAR~\cite{tian2025var} uses a much denser upsampling schedule.

\begin{table}[t]
\setlength{\tabcolsep}{3pt}
\small
  \begin{tabular}{lccc}
    \toprule
    \ourAR & FID$\downarrow$ & KID($\times10^{-3}$)$\downarrow$  & CLIP-score$\uparrow$  \\
    \midrule
    Depth-first & 67.31 & 9.33 & 0.29 \\
    Next scale prediction~\cite{tian2025var}  & 198.37 & 15.42 & 0.21 \\
    \midrule
    Ours (Breadth-first)  & \textbf{56.88} & \textbf{5.79} & \textbf{0.34}\\
  \bottomrule
\end{tabular}
\caption{\textbf{Ablation study on octree node ordering}
}
\lbltbl{abl_order}
\end{table}

\begin{figure*}[t]
    \centering
    \includegraphics[width=\linewidth]{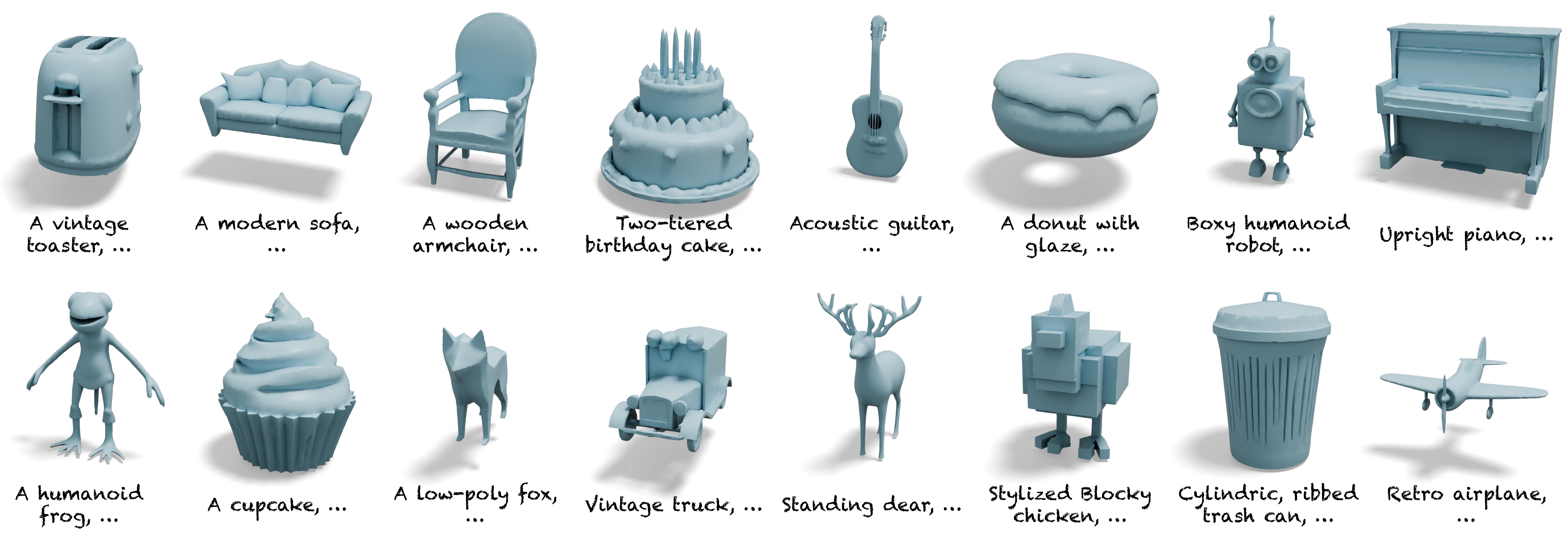}
    \caption{\textbf{Additional shape generation results}}
    \lblfig{gen_appendix}
\end{figure*}

\begin{table*}
\centering
  \begin{tabular}{lccc}
    \toprule
    Method & Token number & Quadric Error Threshold $T$ & Octree Max Depth $L$ \\
    \midrule
    \multirow{4}{*}{Ours} & 266 & 0.001 & 6 \\
    & 439 & 0.0005 & 6\\
    & 625 & 0.0003 & 6\\
    & 1284 & 0.0001 & 6 \\
    \midrule
    \multirow{3}{*}{Ours w/o Adaptive Subdivision} & 148 & - & 4 \\
    & 607 & - & 5 \\
    & 1726 & - & 6 \\
    \bottomrule
\end{tabular}
\caption{\textbf{Hyperparameters.}
}
\lbltbl{hyper}
\end{table*}

\myparagraph{Ablation study on architecture design.} We conduct an ablation study using different pooling operations (conv, concatenation + MLP v.s. our average pooling) in our encoder and removing the hierarchical octree structure (using only leaf nodes). 
\reftbl{abl_arch} shows that different pooling layers result in similar performance. %
In contrast, removing hierarchy, despite comparable reconstruction, degrades downstream generation quality. This shows that our hierarchical tokenization provides a beneficial coarse-to-fine sequence for generation. 

\begin{table}[h]
\setlength{\tabcolsep}{4pt}
\centering
\small
\begin{tabular}{l|cc | cc}
\toprule
  Method & IoU $\uparrow$ & CD ($10^{-3}$) $\downarrow$ & KID ($10^{-3}$) $\downarrow$ & CLIP $\uparrow$  \\
\midrule
Convolution & 88.5 & \textbf{1.75} & 5.81 & \textbf{0.34} \\
Concat + MLP & \textbf{88.6} & 1.82 & 5.95 & 0.33 \\
\midrule
No hierarchy & 88.4 & 1.93 & 13.65 & 0.24\\
\midrule
Ours  & \textbf{88.6} & 1.78 & \textbf{5.79} & \textbf{0.34} \\

\bottomrule
\end{tabular}
\caption{\textbf{Ablation study on architecture design.}}
\lbltbl{abl_arch}
\end{table}

\myparagraph{Comparison with concurrent work.} 
Trellis~\cite{xiang2024trellis} represents 3D objects using fixed-resolution sparse voxel grids, while OctGPT~\cite{wei2025octgpt} uses a non-adaptive octree structure. Unlike ours, both disregard inherent variations in complexity across 3D data, resulting in inefficient tokenization ($\ge$20K tokens). Note that Trellis models both geometry and appearance, which inherently requires a greater number of tokens than geometry-only approaches like OctGPT and ours.
Dora~\cite{chen2024dora} shares our insight about adaptive representational capacity, using a sharp edge sampling strategy for its input point cloud, which could potentially be integrated with our adaptive tokenization. \reffig{recon_rebuttal} shows a comparison of \ours with Trellis and Dora. For Trellis, we use the rendered textured images as input, but only show the geometry reconstruction here for comparison. For Dora, we use the highest-resolution version, which takes a 64K point cloud as input and represents a shape with 4096 tokens. We also report quantitative comparison in \reftbl{recon_appendix}. Our method achieves comparable reconstruction with far fewer tokens than Trellis and Dora (4096 tokens). 

\begin{table}[t]
\small
\centering
\begin{tabular}{l|cc}
\toprule
   & Avg. Token  & CD ($10^{-3}$) $\downarrow$ \\
  \midrule
Trellis~\cite{xiang2024trellis}  & $\sim$20000 & 8.1 \\
\midrule
\multirow{3}{*}{Dora~\cite{chen2024dora}} & 4096 & \textbf{5.2} \\
 & 2048 & 7.3 \\
 & 1024 & 9.4 \\
\midrule
Ours  & 1372 & 7.9 \\

\bottomrule
\end{tabular}
\caption{\textbf{Reconstruction comparison with concurrent work.}}
\lbltbl{recon_appendix}
\end{table}

\begin{figure*}[!h]
    \centering
    \includegraphics[width=.8\linewidth]{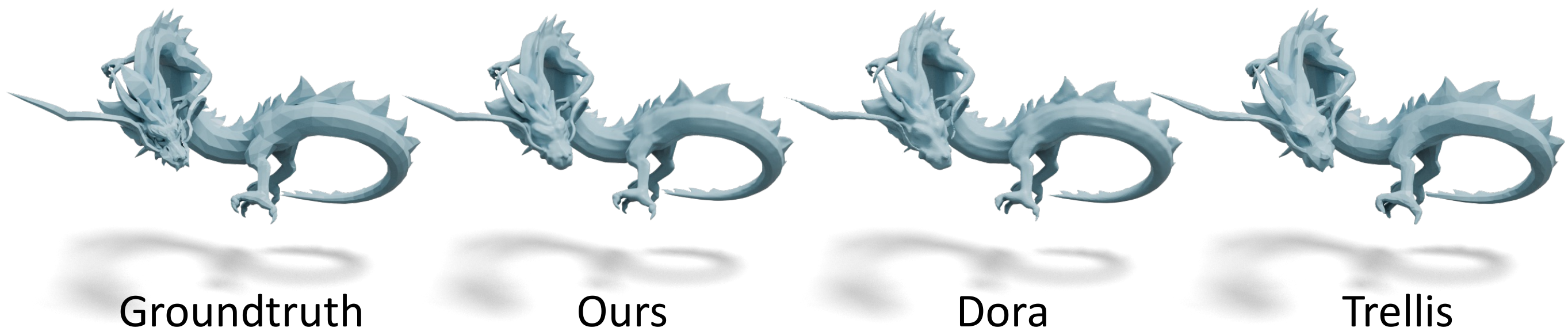}
     \caption{\textbf{Reconstruction comparison with concurrent work}}
    \lblfig{recon_rebuttal}
\end{figure*}

\section{Additional Discussion}

\myparagraph{Societal Impact.} Our Octree-based Adaptive Shape Tokenization approach offers significant potential for advancing 3D content creation. By dynamically allocating representation capacity based on shape complexity, our method substantially reduces computational requirements while maintaining or improving quality. This efficiency enables more detailed and diverse 3D content generation with fewer resources, making high-quality 3D asset creation more accessible and environmentally sustainable. The reduced token count and improved generation capabilities could accelerate applications across gaming, simulation, virtual environments, and digital twins. However, like other generative technologies, our method could potentially be misused to create misleading content. While current human perception can generally distinguish synthetic 3D objects from real ones, we encourage ongoing research into detection methods as these technologies continue to advance.

\end{document}